\documentclass[11pt]{article}

\usepackage{textcomp}  
\usepackage{scalerel}  
\def\mammoth{\scalerel*{\includegraphics{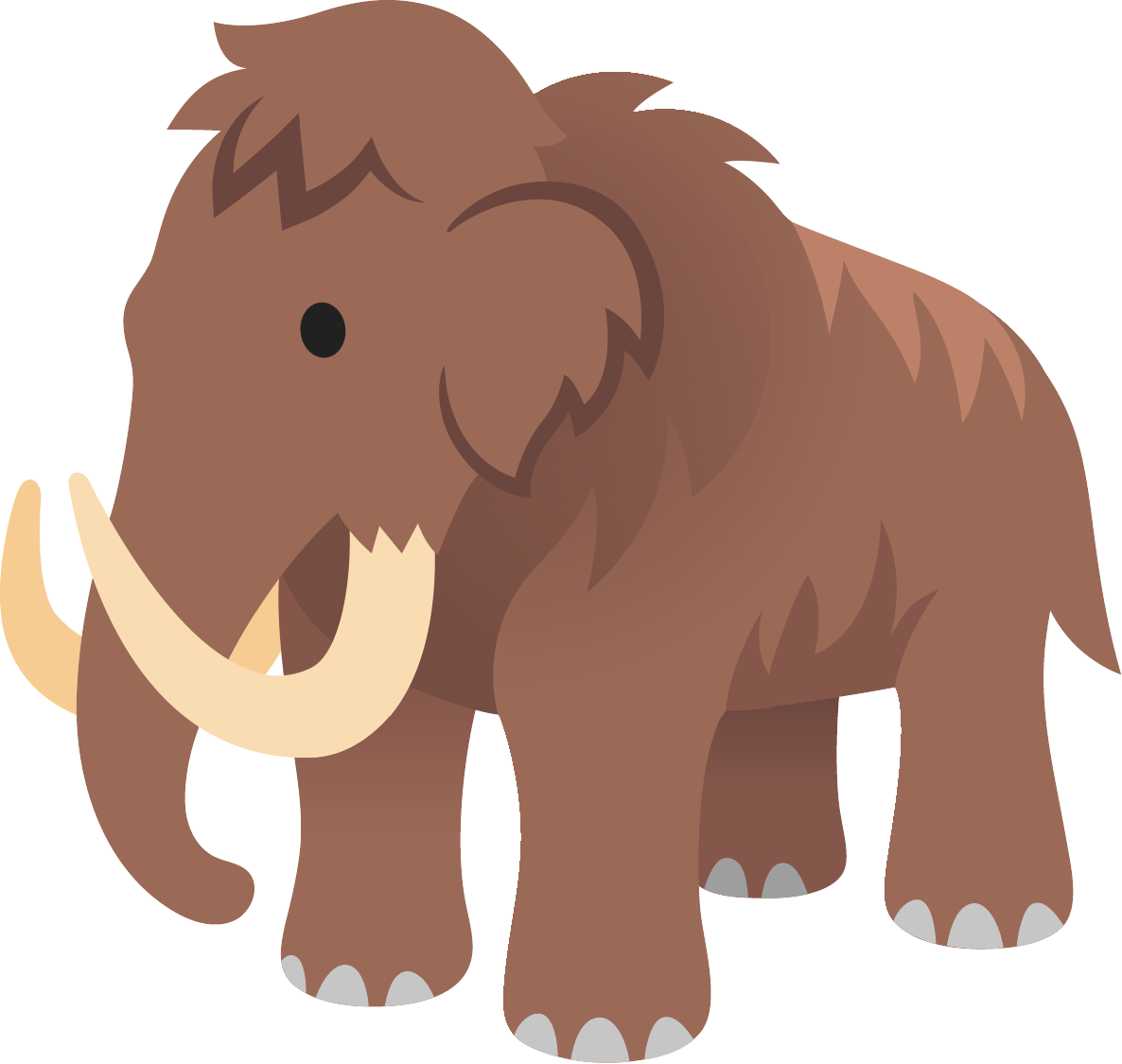}}{\textrm{\textbigcircle}}}
\def\elephant{\scalerel*{\includegraphics{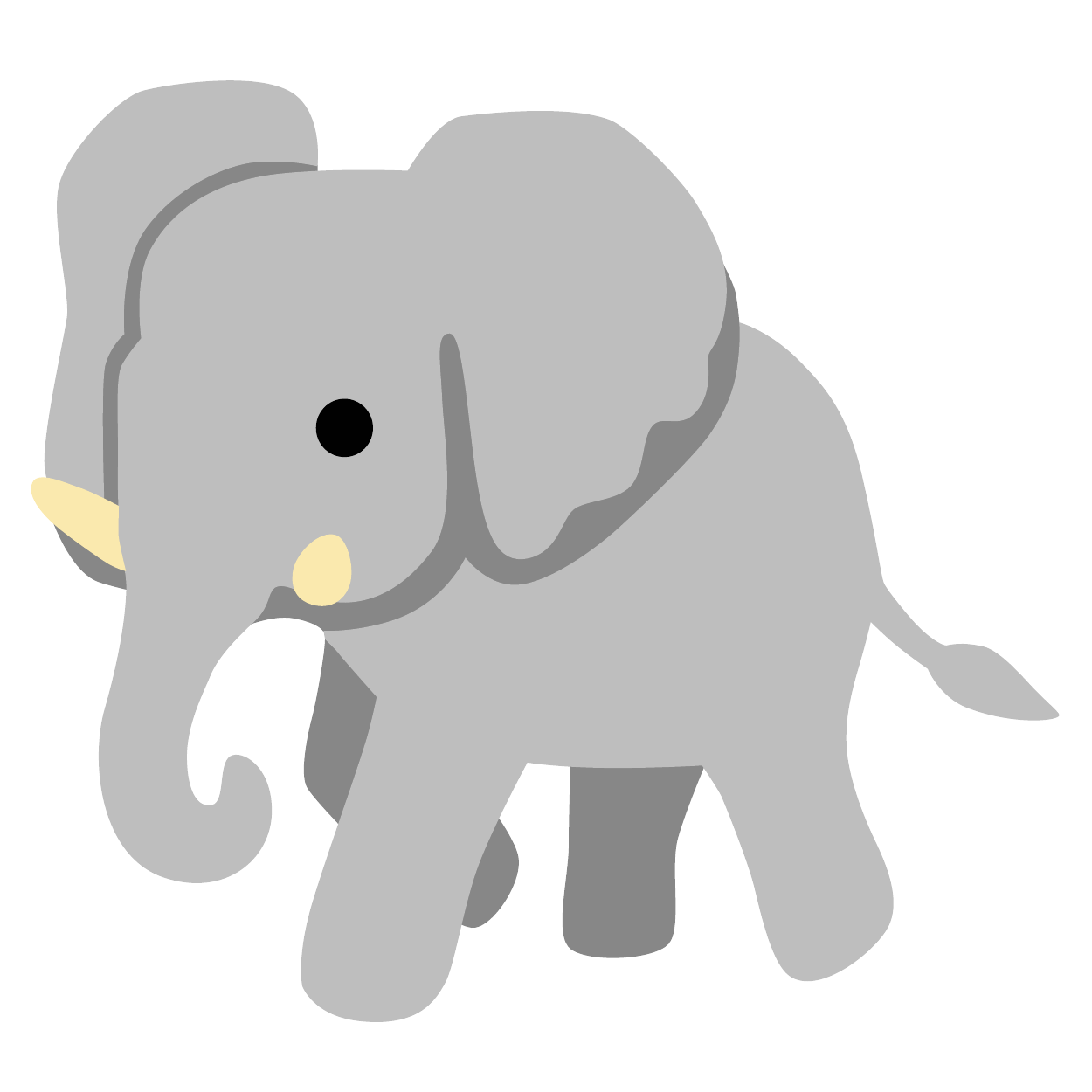}}{\textrm{\textbigcircle}}}
\def\sauropod{\scalerel*{\includegraphics{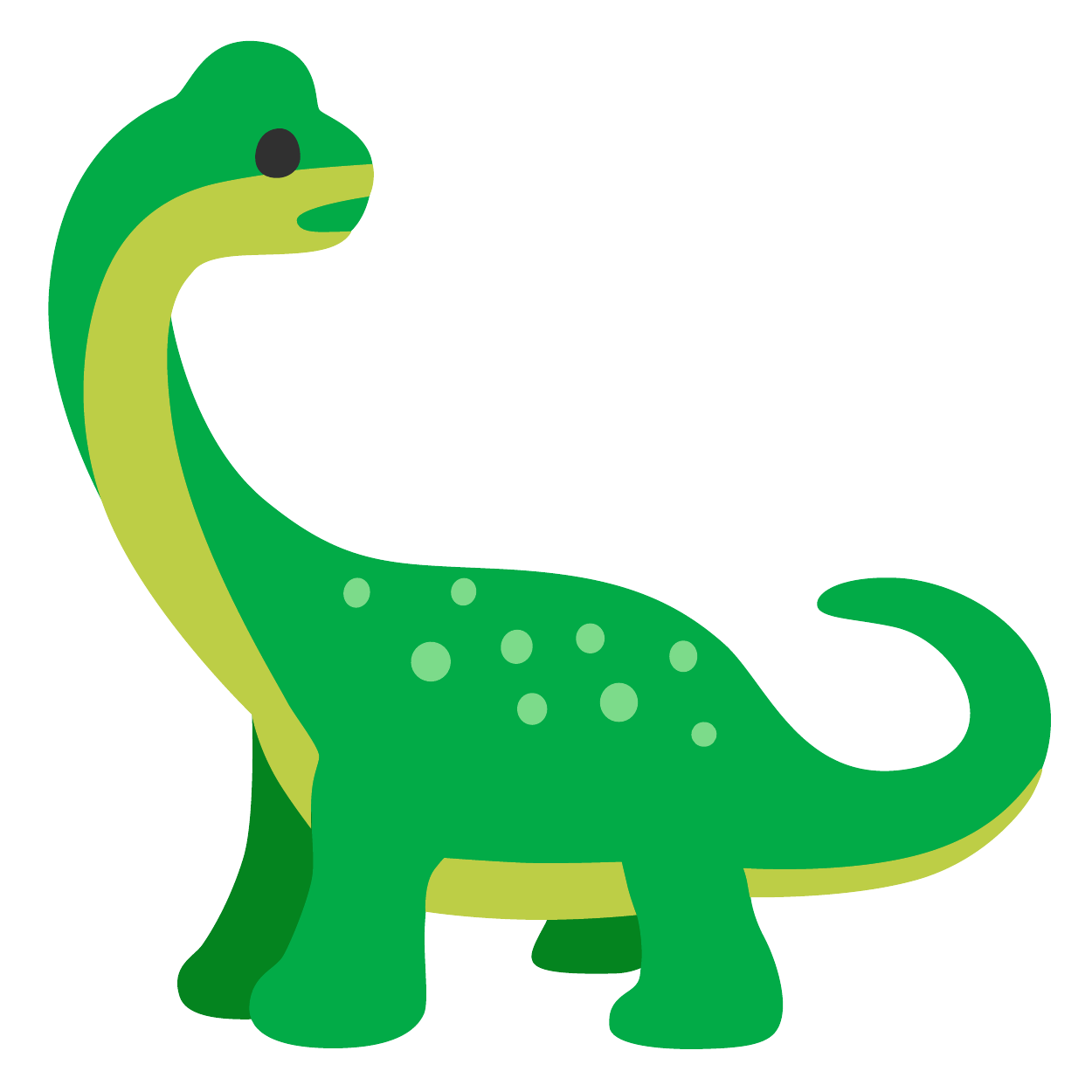}}{\textrm{\textbigcircle}}}
\def\trex{\scalerel*{\includegraphics{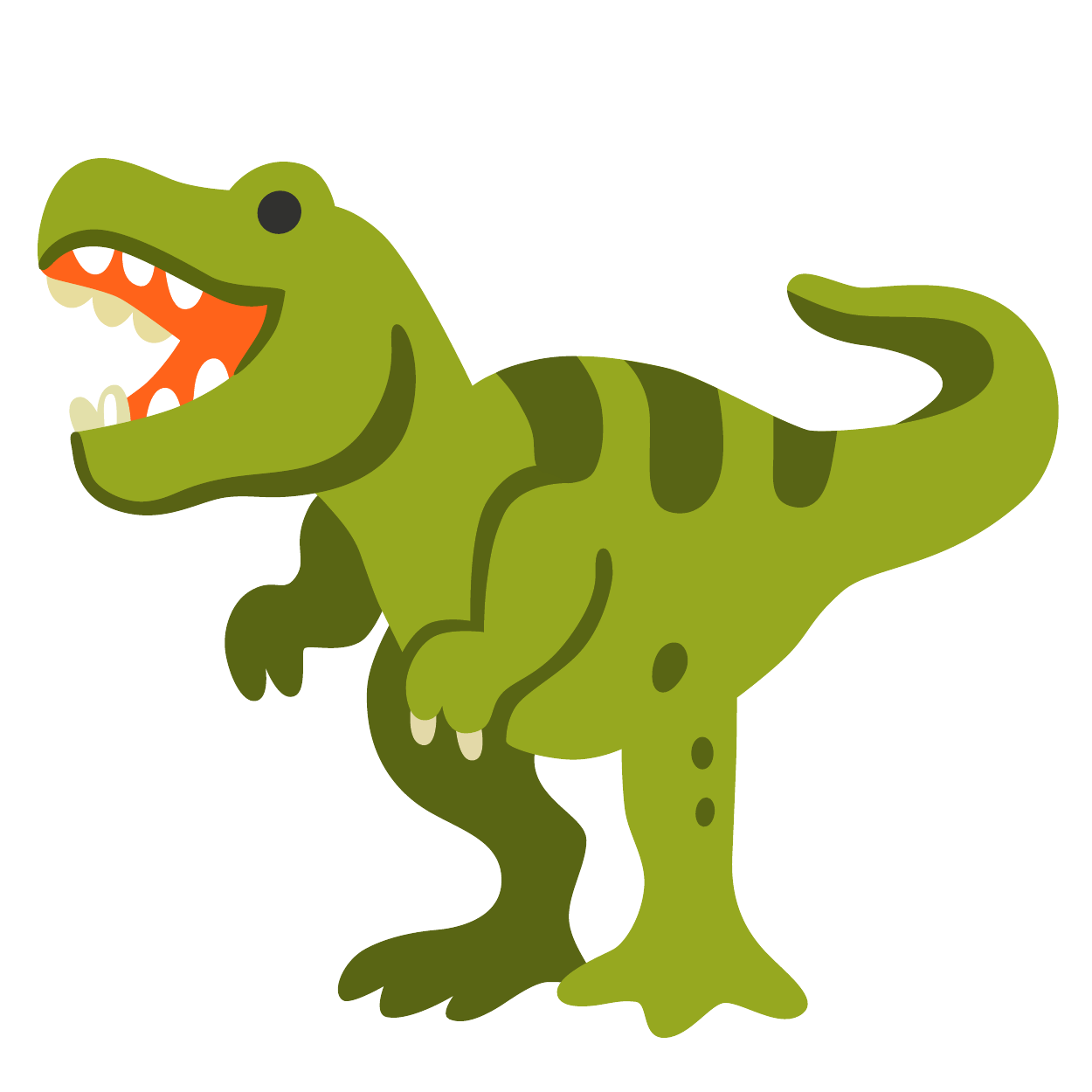}}{\textrm{\textbigcircle}}}
\def\snowflake{\scalerel*{\includegraphics{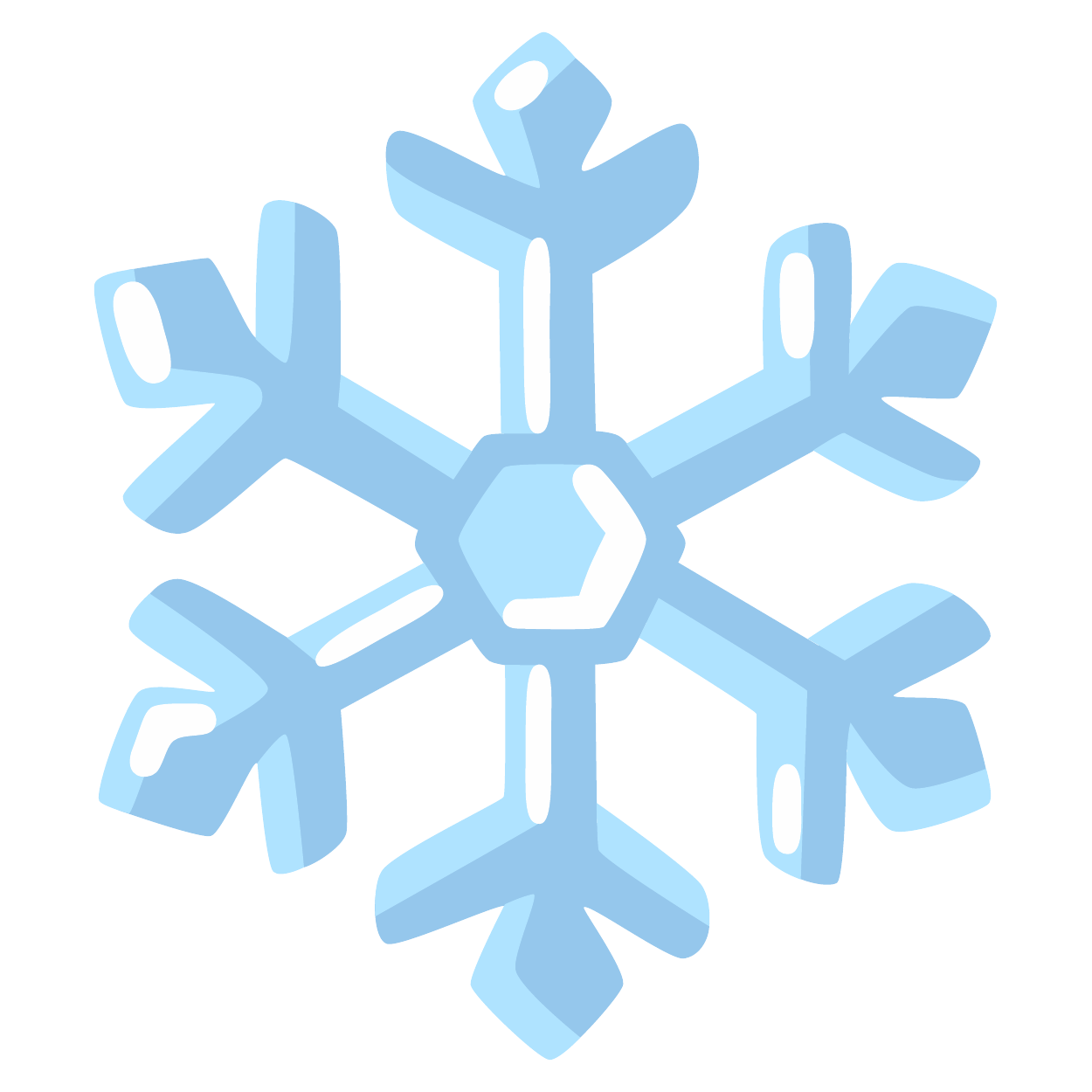}}{\textrm{\textbigcircle}}}
\def\evergreen{\scalerel*{\includegraphics{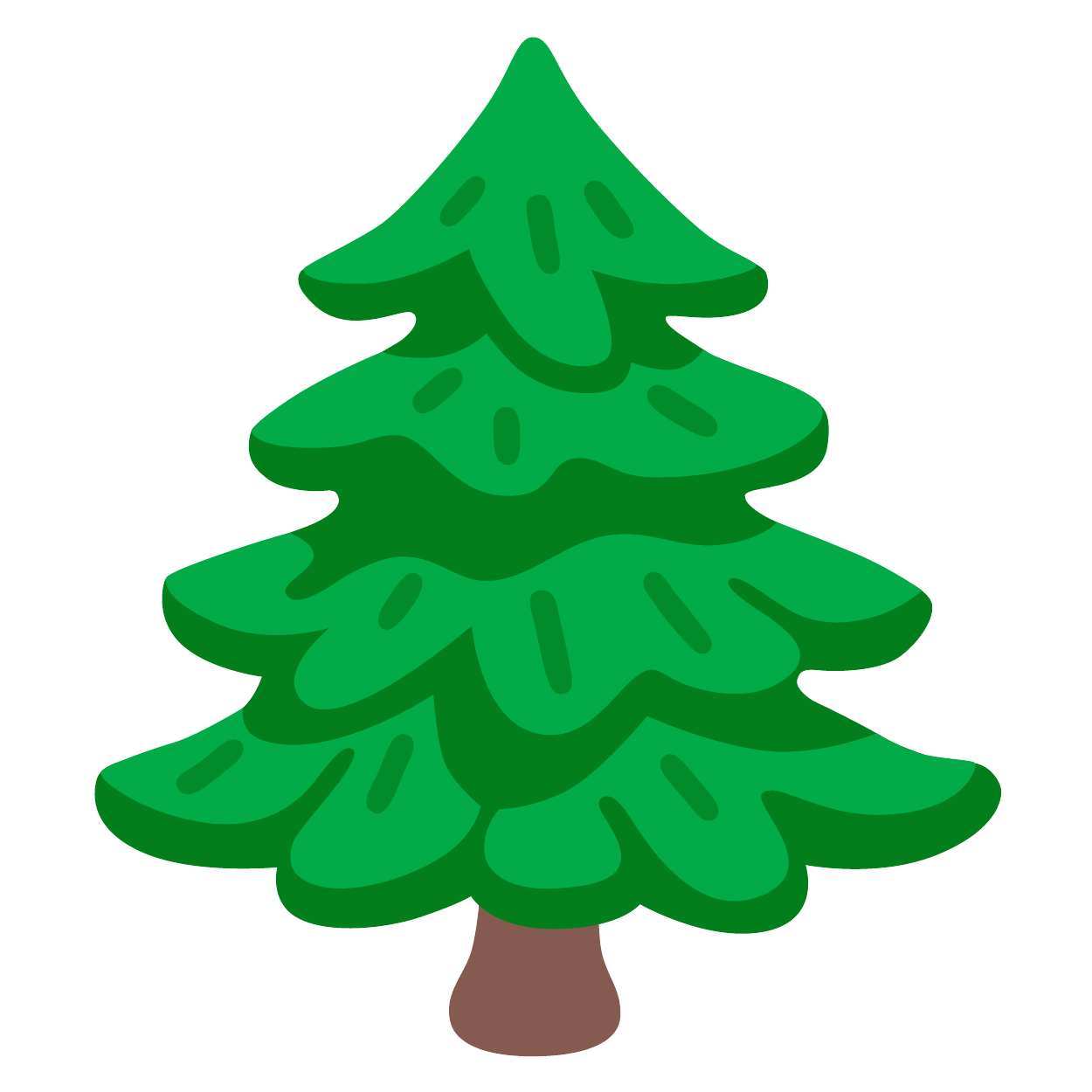}}{\textrm{\textbigcircle}}}
\usepackage{tikz-qtree}
\usepackage{enumerate}
\usepackage{acl}
\usepackage{subfig}
\usepackage{xcolor}
\usepackage{pdfpages}

\usepackage{algorithm}
\usepackage{algpseudocode}
\usepackage{amsmath}
\usepackage{cleveref}

\usepackage{ragged2e}
\usepackage{times}
\usepackage{latexsym}

\usepackage[T1]{fontenc}

\usepackage[utf8]{inputenc}

\usepackage{microtype}

\usepackage{inconsolata}

\title{MAMMOTH ~ \mammoth\\ Massively Multilingual Modular Open Translation @ Helsinki}

\author{Timothee Mickus\textsuperscript{\mammoth\evergreen} \quad 
    Stig-Arne Grönroos\textsuperscript{\mammoth\sauropod\evergreen} \quad 
    Joseph Attieh\textsuperscript{\mammoth\snowflake} \quad
    Michele Boggia\textsuperscript{\snowflake} \\
    \textbf{Ona De Gibert}\textsuperscript{\mammoth\snowflake} \hfill
    \textbf{Shaoxiong Ji}\textsuperscript{\mammoth\snowflake} \hfill 
    \textbf{Niki Andreas Loppi}\textsuperscript{\trex\snowflake}\\ 
    \textbf{Alessandro Raganato}\textsuperscript{\elephant\snowflake} \hfill 
    \textbf{Raúl Vázquez}\textsuperscript{\mammoth\snowflake} \hfill
    \textbf{Jörg Tiedemann}\textsuperscript{\mammoth}
    \\[0.2cm] 
    \qquad\qquad\qquad\qquad\qquad\textsuperscript{\mammoth}University of Helsinki \hfill \textsuperscript{\sauropod}Silo AI     \qquad\qquad\qquad\qquad\qquad\null \\ 
    \qquad\qquad\qquad\qquad\qquad\textsuperscript{\elephant}University of Milano-Bicocca \hfill \textsuperscript{\trex}NVIDIA\qquad\qquad\qquad\qquad\qquad\null  \\[0.2cm] 
    \qquad\qquad\qquad\qquad\qquad\textsuperscript{\evergreen}Equal contribution\hfill\textsuperscript{\snowflake}Alphabetical order\qquad\qquad\qquad\qquad\qquad\null}

\begin{document}
\maketitle
\begin{abstract}
NLP in the age of monolithic large language models is approaching its limits in terms of size and information that can be handled. 
The trend goes to modularization, a necessary step into the direction of designing smaller sub-networks and components with specialized functionality. 
In this paper, we present the MAMMOTH toolkit: a framework designed for training massively multilingual modular machine translation systems at scale, initially derived from OpenNMT-py and then adapted to ensure efficient training across computation clusters.
We showcase its efficiency across clusters of A100 and V100 NVIDIA GPUs, and discuss our design philosophy and plans for future information.
The toolkit is publicly available online.

\begin{figure}[ht]
\centering
\begin{minipage}[c]{0.075\linewidth}
\centering
\includegraphics[width=1\textwidth]{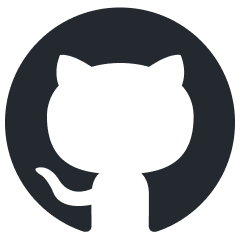}
\end{minipage}
\begin{minipage}[c]{0.775\linewidth}
\small ~ \href{https://github.com/Helsinki-NLP/mammoth}{\tt github.com/Helsinki-NLP/mammoth}
\end{minipage}%
\end{figure}

\end{abstract}

\section{Introduction}
The field of NLP has recently witnessed a hastened transition towards ever-larger monolithic neural networks, 
exposed to gargantuan amounts of data so as to properly fit their humongous number of parameters.
There is also a growing consensus that this is not a sustainable trend: 
This approach falters whenever data is scarce, and leads to costs---both financial and ecological---that cannot be disregarded.

The problems of scalability are especially prominent in the field of multilingual NLP. 
Scaling a multilingual model to a high number of languages is prone to suffer from interference, also known as the curse of multilinguality, leading to degradation in per-language performance, mainly due to the limited model capacity \citep{conneau-etal-2020-unsupervised,wang-etal-2020-negative}. 
Increasing the overall model size, on the other hand, hits the ceiling in terms of trainability limited by hardware, data and training algorithms.
Modularity is one approach attempting to answer the challenges of scalability.

\paragraph{What is modularity?} 
Modularity can be viewed in two complementary ways: as sparsity or as conditional computation.
In the former, modularity enforces a principled sparsity of the network, so as to allow a model to be large at training time, but small during inference. 
In the latter, a modular approach entails routing the information flow to a module in order to select specific model parameters to be used in specific circumstances.

A canonical example is the use of language-specific encoders and decoders for machine translation:
For the translation direction going from a source language $L_S$ to a target language $L_T$, one would use an encoder trained on on the data from the $L_S$ source language, and a decoder trained to handle any and all inference to the $L_T$ target language.
Note, that the encoder will typically have seen data with $L_S$ as source language but other languages than $L_T$ as target languages, and mutatis mutandis for the decoder.
This dynamic selection of modules entails a principled sparse activation: 
In this translation scenario, any encoder or decoder linked to some third language $L_Z$ would not contribute to the computation, as if was set at zero.

This design can therefore lead to more efficient inference, since we can avoid computations for a large proportion of the parameters.
The modularity also aids in the interpretability of parameters, as it is easy to determine which tasks a parameter contributes to.
It also fosters the design of reusable neural network components: 
Modules can in principle be combined to allow for zero-shot adaptations to novel tasks and situations \citep{pfeiffer-etal-2021-adapterfusion}.

\paragraph{What we provide.} 
One of the challenges that come with the study of modularity is the lack of a consensual, broadly available framework for designing and handling such models. 
The MAMMOTH toolkit is meant to address this gap in the current NLP ecosystem.
We build upon the OpenNMT-py library \citep{klein2018opennmt} and provide a set of utilities to effectively train modular encoder-decoder systems. 
The MAMMOTH toolkit is tailored toward efficient computation across clusters of compute nodes, and covers a broad range of architectures and use-cases.
We also inherit the user-centric concerns of \citet{klein2018opennmt}:
The design of MAMMOTH include \emph{transforms}, externalized computation steps providing users with means to include arbitrary preprocessing to better suit their experimental needs; moreover we provide utilities to assist users with designing configuration files for massive and complex experiments semi-automatically.
The MAMMOTH toolkit is available publicly under a CC-BY license and warmly welcomes prospective developers and researchers wishing to contribute or request the implementation of specific features. 

\section{Related work}
\label{sec:relatedwork}

While there exist many open-source frameworks for training NMT systems that have been used for experiments in modular NMT, such as fairseq \citep{ott2019fairseq}, to the best of our knowledge none of them is specifically targeted at modularity. MAMMOTH is the first open-source toolkit to jointly address the issues of scalability, multilinguality and modularity.

The most relevant point of comparison would be the AdapterHub of \citet{pfeiffer2020adapterhub}: 
It extends the \texttt{transformers} library (\citealp{wolf2020huggingfaces}; an NLP-centric model sharing platform and training library) so as to enable training of adapter modules for pre-trained state-of-the-art systems.
We base our MAMMOTH modular toolkit on the widely used OpenNMT-py \citep{klein2018opennmt}, a customizable library for training NMT and NLG models with a focus on efficiency based on PyTorch: 
The more thorough documentation and more systematic organization of OpenNMT-py proved a better starting point for MAMMOTH than the fairseq or transformers libraries.

Another motivation behind MAMMOTH is the lack of a standard for the different architectures of existing works on modular systems. 
Modularity in multilingual NMT has been addressed in a wide range of custom implementations. 
One common approach is to train language-specific encoders and decoders.
\citet{vazquez-etal-2020-systematic} introduce the attention bridge, an intermediate cross-lingual shared layer in between the encoder and decoder. 
Similarly, \citet{escolano-etal-2021-multilingual} train language-specific encoders-decoders without sharing any parameters at all. 
\citet{purason-tattar-2022-multilingual} also exploit this method and explore different sharing strategies for the decoder while keeping the decoder language or language-group-specific. 
\citet{yuan2023lego} experiment with massive multilingual MT and propose a plug-and-play approach with detachable modules per language. 
Other methods investigate the use of language-specific transformer layers \citep{pires-etal-2023-learning}, by keeping some layers source or target language-specific in the encoder. 
MAMMOTH supports all of the above, and provides a new unifying standard framework for training and testing modular NMT at scale.

\section{Toolkit design}
\label{sec:description}

\begin{figure*}[th]
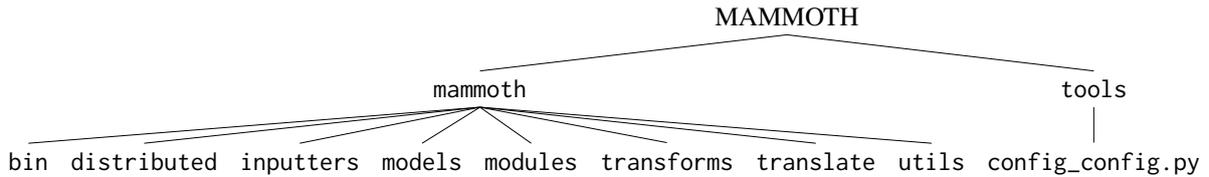

    \centering
    \resizebox{\linewidth}{!}{
    \Tree [ [ \texttt{bin} \texttt{distributed} \texttt{inputters} \texttt{models} \texttt{modules} \texttt{transforms} \texttt{translate} \texttt{utils} ].\texttt{mammoth} [ \texttt{config\_config.py} ].\texttt{tools} ] .MAMMOTH
  }
    \caption{Code repository structure overview}
    \label{fig:overview}
\end{figure*}

We now turn to a description of our toolkit. 
This section first details the requirements for the toolkit
\Cref{sec:description:requirements}, before moving to the 
design philosophy we adopted in response to the practical needs outlined in \Cref{sec:description:design-principles}.
Finally, we list and describe all major components of our toolkit in \Cref{sec:description:structure}.

\subsection{Required functionalities}
\label{sec:description:requirements}

\paragraph{Broad architecture coverage.}
We aim at a toolkit that covers the major modular architectures proposed in related work (see \Cref{sec:relatedwork}) in order to allow systematic studies over a range of implementations in a single all-encompassing framework, ruling out cross-framework variation that prevents a fair comparison of approaches.

In practice, the MAMMOTH toolkit focuses on architectures where modular components can be defined \textsl{a priori} and operate as \textsl{separable units}. The goal is to allow efficient training of large-scale modular systems from scratch with the possibility of a flexible asymmetric distribution of components across large compute clusters. Furthermore, we aim for efficient inference with a decomposable network where only necessary components need to be loaded for specific tasks.
%
This leaves out sub-network selection approaches, such as the Language Specific Sub-network architecture \citep[LASS;][]{lin-etal-2021-learning} where a unified multilingual network is split down into language-specific parameters.
This also removes dynamic routing approaches, such as the Mixture-of-Experts \citep[MoE;][]{shazeer2017outrageously} architecture, where a model learns a gating function that activates modules in the network.
In such architectures, it is not possible to determine in advance which parameters will be active---and thus all parameters need to be loaded into memory. 
In contrast, the component-level modularity used in MAMMOTH allows loading only the necessary parameters.

\paragraph{Efficient Training.}
An important challenge that modular approaches are faced with is owed to their principled sparsity: 
It is not necessarily feasible---and most often not desirable---to host a copy of all modules on every available computation device. 
As such, modules have to be assigned to specific devices, which in turn entails that a massively modular system must also be able to handle communication between any two devices that both host a copy of the same module.
Minimizing the necessary communication across devices through optimal device assignments of modules becomes a critical aspect for efficient modular training.
A practical reality of multilingual and multitask settings is that often data is not equally available for all languages and tasks. 
Natively handling skewed datasets and ensuring that no computation device is ever idle is also crucial to efficiency.

\subsection{Design principles}
\label{sec:description:design-principles}

The MAMMOTH toolkit is designed around the concept of a \emph{task}.
A task is the conjunction of three elements, kept constant during the whole of training: 
\begin{enumerate}[(i)]
    \item a set of modules;
    \item a set of preprocessing steps; and
    \item a single dataset (typically a parallel corpus).
\end{enumerate}
In short, a task corresponds to a specific model behavior.
In translation settings, a task will therefore correspond to a specific translation direction (say translating from Swahili to Catalan):
All training datapoints for this direction  (i) must involve the same modules (pertaining to Swahili encoding and Catalan decoding); (ii) must be preprocessed with the same tokenizers; and (iii) can be grouped into a single bitext.

We furthermore enforce that a task is tied to a specific compute node and device; in other words, we associate each task with an available GPU, and host a copy of all modules relevant to that task on said GPU.
This greatly simplifies questions of device allocation and communication efficiency, since we can examine how allocating specific tasks to specific GPUs will impact communication across devices. 
In short, we can aim to minimize communication by associating tasks that involve the same components on the same computation device, thereby limiting the number of module copies for which gradient would need to be synchronized.

\begin{figure*}[ht!]
\centering
\subfloat[Example configuration for task-specific encoders and decoders]{
    \resizebox{0.3\linewidth}{!}{
        \includegraphics[trim={4cm 13cm 10cm 4.5cm},clip]{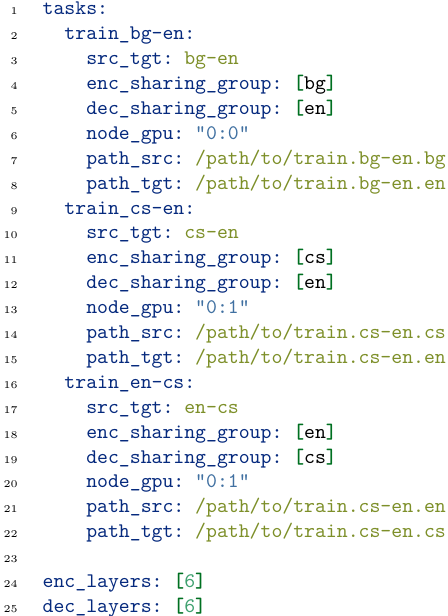}
    }
}
\hfill
\subfloat[\label{cfg:arb}Example configuration for arbitrarily shared layers in encoders, and task-specific decoders]{
    \resizebox{0.3\linewidth}{!}{
        \includegraphics[trim={4cm 13cm 10cm 4.5cm},clip]{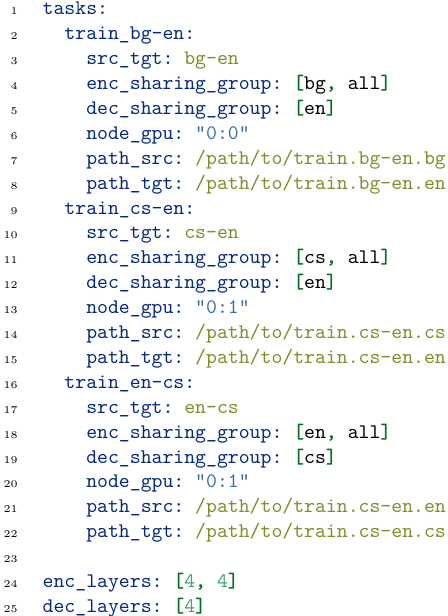}
    }
}
\hfill
\subfloat[Example configuration for a non-modular multilingual system. A lack of target language specific parameters necessitates adding a language  token using a prefix transform.]{
    \resizebox{0.27\linewidth}{!}{
        \includegraphics[trim={4cm 10.3cm 10cm 4.5cm},clip]{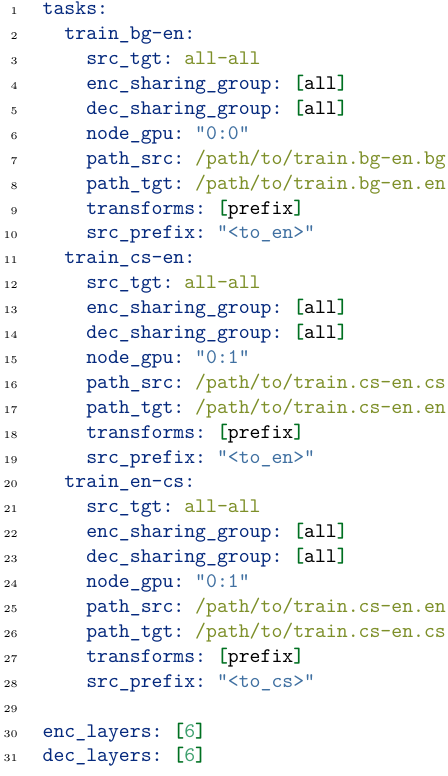}
    }
}

    \caption{Snippets from example configurations}
    \label{fig:example-confs}
\end{figure*}

In \Cref{fig:example-confs}, we showcase a few examples of configurations file snippets to illustrate how tasks can be defined.
In essence, the configuration file expects the \texttt{tasks} key to be a mapping of task identifiers (e.g., ``\texttt{train\_bg-en}'') to task definitions.
Each task definition must explicitly state the sequence of modules to be used for the encoders and decoders (or ``sharing groups'').
This allows for a highly flexible modular configuration, ranging from systems that only involve task-specific modules to non-modular systems.
The current main limitations we impose are that (i) all tasks must involve the same number of modules,\footnote{
    Note that this does not entail that all tasks involve the same degree of sharing: 
    By setting up task-specific modular components, one can easily define a pipeline that formally contains the same number of modules (so as to satisfy this requirement) and that is only applied to a given task. 
    To take a concrete example, we could define an NMT system where encoders would be comprised of one language-specific module and one language family-specific module: Isolated languages such as Basque would therefore not share any of their encoder parameters with any other languages.
} and (ii) module definitions are specific to a given sequential position. 
In other words, defining two tasks with encoder sharing groups \texttt{[x, y]} and \texttt{[y, x]} entails defining four modules, not two.
These limitations however allow us to factor out the number of layers each module has, by means of the \texttt{enc\_layers} and \texttt{dec\_layers} keys.

\subsection{Implementation}
\label{sec:description:structure}

Our code-base is historically based on the OpenNMT-py library \citep{klein2018opennmt}. 
Although the changes accrued to convert it to a modular framework have proven significant enough to require deep refactoring and restructuring, readers may find referring to \citet{klein2018opennmt} a useful addition to the present description to cover the basic aspects of the NMT toolkit.
Some practical implementation choices we inherit from OpenNMT-py include the fact that our framework is based on PyTorch \citep{NEURIPS2019_bdbca288} as well as the presence of \emph{transforms} and YAML-based configuration files.

\Cref{fig:overview} provides an overview of the major elements included in the MAMMOTH toolkit. 
Utilities listed in the \texttt{tools} are intended to facilitate setting up training, conducting experiments, or evaluating models.
The core source-code itself is filed under the \texttt{mammoth} directory.
The source files are grouped in eight python submodules.

\paragraph{The \texttt{bin} submodule.} 
The \texttt{bin} submodule contains utilities for training modular systems and translating using MAMMOTH systems.

\paragraph{The \texttt{transforms} submodule.} 
The \texttt{transforms} submodule contains a list of default transforms that toolkit users can easily expand to suit their preprocessing needs.
Currently, the MAMMOTH toolkit contains transforms for:
    external tokenization (e.g., using SentencePiece, \citealp{kudo-richardson-2018-sentencepiece}) and subword regularization \citep{kudo-2018-subword};
    denoising auto-encoding task, using a BART-style \citep{lewis-etal-2020-bart} or a MASS-style \citep{pmlr-v97-song19d} objective;
    filtering low-quality parallel sentences on the fly;
    and injecting prefixes, such as control and language tokens or decoder-side prompts.

\begin{algorithm}[ht]
\small
\caption{Gradient accumulation and communication for one datapoint of task $\mathcal{T}$}\label{alg:grad-accum}
\begin{algorithmic}
\Require set of modules in the complete model, \\ C = $\left\{ C_1, \dots, C_u \right\}$
\Require set of modules on the device, \\ D = $\left\{ D_1, \dots, D_v \right\} \subseteq C$
\Require ordered set of modules used in task $\mathcal{T}$, \\ $T = \left\{T_1, \dots, T_w\right\} \subseteq D$
\Require inputs and labels, $\langle\mathbf{x}, \mathbf{y}\rangle \in \mathcal{D}_\mathcal{T}$
\State \null
\State \null \Comment{\textit{forward pass}}
\State $\mathbf{h}_0 \gets \mathbf{x}$
\For{$i \in \{1, \dots, w \}$} 
    \State $\mathbf{h}_i \gets T_{i}\left(\mathbf{h}_{i-1}\right)$
\EndFor
\State \null
\State \null \Comment{\textit{communicate modules used}}
\State $\mathrm{ready}_T \gets \left( \mathbf{1}_T(D_1), ~ \dots,  ~ \mathbf{1}_T(D_v)  \right)$
\State $\mathbf{n} \gets$ broadcast $\mathrm{ready}_T$, reduce by sum
\State \null
\State \null \Comment{\textit{backward pass}}
\For{$j \in \{1, \dots, v \}$} 
    \State $\Delta D_{j} \gets\begin{cases}
    - \nabla_{\mathbf{h}_i} D_{j} & D_{j} \in T \\
    0 & D_{j} \notin T \\
    \end{cases}$
    \State $\Delta D_{j} \gets$ broadcast $\Delta D_{j}$, reduce by sum
    \State $\Delta D_{j} \gets \Delta D_{j} / \mathbf{n}_j$
\EndFor
\end{algorithmic}
\end{algorithm}

\paragraph{The \texttt{distributed} submodule.} 
The \texttt{distributed} submodule contains the core implementation of the conceptual tasks we outlined in \Cref{sec:description:design-principles}.
It also defines routines for efficient communication in modular settings. 
In particular, we define an algorithm for broadcasting and synchronizing gradients across all tasks shown in \Cref{alg:grad-accum}:
The gist of it is that we
are able to use the a priori structure to omit communication of parameters that are not on the device at all.
For the modules on the device, we signal in how many tasks it has been used, and therefore has a gradient, and rely on this information to sum and renormalize gradients appropriately.
The \texttt{distributed} submodule also contains explicit implementations for computation contexts (distributed vs. single-GPU vs. CPU-bound training), as well as logic to handle uneven task distributions.

\paragraph{The \texttt{inputters} submodule.} 
The \texttt{inputters} submodule contains all code logic for handling data: 
It provides objects for representing parallel corpora, handling batching through reservoir sampling, and multiplexing batch streams from multiple tasks whenever more than one task is allocated to a given device.

\paragraph{The \texttt{modules} and \texttt{models} submodule.}
These two python submodules contain code logic for defining specific PyTorch components and grouping them together in coherent models.

\paragraph{The \texttt{translate} submodule.}
The \texttt{translate} submodule contains all code logic for handling inference.

\paragraph{The \texttt{utils} submodule.}
The \texttt{utils} submodule regroups all remaining code logic, including optimizers, loss computation, early stopping and tensorboard reporting.

\paragraph{The \texttt{config-config} tool.}
We prefer explicit over implicit when it comes to the configuration yaml file, even though the configuration can become verbose and repetitive. The \texttt{config-config} tool makes configuring MAMMOTH more user friendly.
When given a simple meta-configuration file (see \Cref{adx:metaconfig} for an example), it can
perform the following operations to generate the complete configuration:
\begin{itemize}
    \item Path templating for massively multilingual corpora with various directory structures (See \Cref{adx:pathtemplate} for details),
    \item Find which tasks have data in the corpus,
    \item Determine task weighting and curriculum,
    \item Determine language groups by clustering,
    \item Configure the layerwise parameter sharing groups of tasks  (See \Cref{adx:parametersharing} for details),
    \item Ensure that the correct \emph{transforms} are set for translation and denoising autoencoder tasks,
    \item Allocate tasks to nodes and GPUs. A local search procedure is used, taking into account parameter sharing groups and tasks delayed by curriculum weighting.
    \item Determine the adapter configuration for each task.
\end{itemize}

\section{Performances}

We utilize the Europarl dataset~\citep{koehn-2005-europarl}  - a multilingual resource extracted from European Parliament proceedings containing texts in 21 European languages - for model training to showcase the efficiency of our toolkit. 
We report performances on the Europarl dataset across various parameter-sharing schemes and computing clusters. 

\paragraph{Modeling.} We use a SentencePiece model trained on OPUS Tatoeba Challenge data with 64k vocabulary size.\footnote{\url{https://object.pouta.csc.fi/Tatoeba-MT-spm/opusTC.mul.64k.spm}}
We adopt three sharing schemes: 1) a language-specific one that uses a balanced architecture with a 6-layer transformer encoder and a 6-layer decoder for each language, 2) a partially shared one that has eight layers of encoders (4 shared and language-specific layers) and four layers of decoders, and 3) a fully shared one with 9-layer encoder and 4-layer decoder for all languages.
For the transformer network, we enable position encoding and use a dimension of 512 and 8 attention heads. The feed-forward dimension within the transformer is 2048. 

\paragraph{Setup.}
We utilize two types of GPUs: NVIDIA V100 and A100. We run the toolkit with Python 3.9.16 and PyTorch 2.1.0. We use a maximum sequence length for source and target languages of 200, and a  batch size of 4096 tokens. 
The detailed setup guide for the experiment is available in our documentation,\footnote{\url{https://helsinki-nlp.github.io/mammoth/examples/train_mammoth_101.html}} including data processing, configurations, and launching scripts. 

\paragraph{Benchmarking results.}


We studied the performance of MAMMOTH on CSC's NVIDIA V100 and A100 clusters, where each node contains four NVIDIA V100/A100 GPUs connected via NVLink and nodes are interconnected via InfiniBand. 
Scaling benchmarks were undertaken such that the number of tasks increases proportional to the number of allocated GPUs as we are interested in the ability to scale to larger problems with similar load per GPU. 

In practice, we train one task on each available GPU for this benchmark.
For benchmarking purposes, this task is defined over synthetic data derived from the Europarl dataset \citep{koehn-2005-europarl}:
We concatenate bi-texts for all available translation directions, and then randomly split it into 20 sub-corpora. 
While individual data-points remain coherent, this shuffling process allows us to sidestep concerns about variation in linguistic factors, such as sentence length, that were found to be a concern in an earlier iteration of this benchmark.

Moreover, we studied task
scaling with three different approaches. 
Firstly, an approach where all source languages use language-specific encoders and decoders, denoted by ``independent.'' 
Secondly, an approach where all source languages use the same shared encoder but decoders are target language specific, denoted by ``shared.'' 
Finally, an approach
denoted as ``partially shared'',
where the encoder begins with a stack of language-specific layers, followed by a stack of shared encoder layers,
finally passing information to language-specific decoders,
similar to the setup discussed in \Cref{cfg:arb}. 

\begin{figure}
  \includegraphics[width=\linewidth]{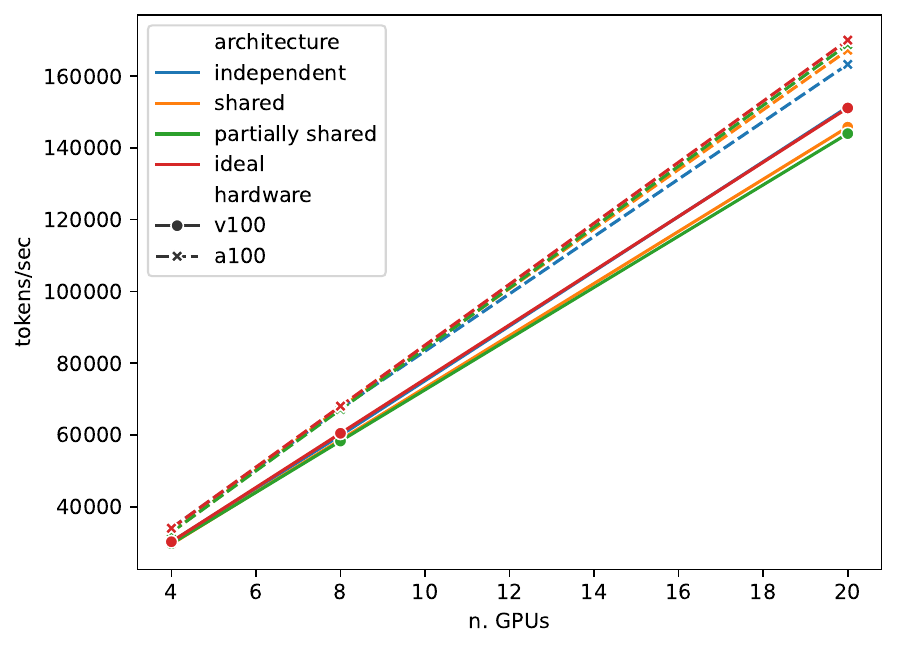}
  \caption{Scaling on V100 and A100 clusters}
  \label{fig:scaling}
\end{figure}

\Cref{fig:scaling} shows task scaling benchmarks from a single node up to 5 nodes using the synthetic dataset. 
We achieve nearly ideal scaling in all of the scenarios: 
Drops in performance compared to our ideal reference are limited to $5\%$ at most. 

To give more context to the benchmarks, we measured the memory footprint and utilizations independently on each GPU. 
The mean utilization percentage with the five nodes run with language-specific encoders and decoders was $85 \%$, and the device memory footprint was $7.7$G out of $32$G. 
Furthermore, we also did performance profiling using NVIDIA Nsight Systems tool to better understand the communication, using the 2-node setup. 
All communication in the code is outsourced to the NVIDIA Collective Communications Library (NCCL) and at present, it appears that approximately 28 \% of the GPU-kernel execution time is spent on NCCL Allreduce. 
Further communication studies and optimizations with different architectures and setups remain subject to future work. 
Nevertheless, as we are able to achieve linear scaling beyond a single node, it suggests that the communication overheads are alleviated at scale (relative to tokens processed per second).


\paragraph{Environmental costs.} We run the benchmarking experiments on CSC's carbon-neutral data center powered by hydropower. 
We measure our carbon footprint using the direct eq. CO2 of 0kg/kWh and indirect of 0.024kg/kWh as the carbon efficiency.\footnote{\url{https://www.syke.fi/fi-FI/Tutkimus__kehittaminen/Kiertotalous/Laskurit/YHiilari}} 
Each benchmarking experiment runs for around one hour. 
Our total carbon footprint is 0.11 kg eq. CO2.
The estimated carbon footprint for training a model over 24 hours is 0.14 kg eq. CO2.

\section{Conclusions and next developments}
In this paper, we have introduced MAMMOTH, a toolkit for training modular encoder-decoder neural networks at scale.
The toolkit is publicly available under a CC-BY license, and we warmly welcome the help of new developers and researchers wishing to extend it.

Further planned development of the MAMMOTH toolkit will specifically focus on the following elements:
\begin{itemize}
    \item Interfacing our toolkit with the popular HuggingFace framework, to allow a wider diffusion of MAMMOTH-based modules and reusing existing foundation models for the initialization of modular systems
    \item Interfacing the MAMMOTH toolkit with the OPUS ecosystem \citep{tiedemann2012parallel}, and in particular the OpusFilter tools \citep{aulamo-etal-2020-opusfilter} so as to delegate data selection to a dedicated third party.
    \item Providing support for partially frozen modular systems, which would enable adapter-style parameter-efficient fine-tuning.
    \item Continuing our work of including modular approaches, in particular continuous prefixes.
\end{itemize}

\section*{Acknowledgments}
\noindent
{ 
\begin{minipage}{0.1\linewidth}
    \vspace{-10pt}
    \raisebox{-0.2\height}{\includegraphics[trim =32mm 55mm 30mm 5mm, clip, scale=0.18]{logos/erc.ai}} \\[0.25cm]
    \raisebox{-0.25\height}{\includegraphics[trim =0mm 5mm 5mm 2mm,clip,scale=0.075]{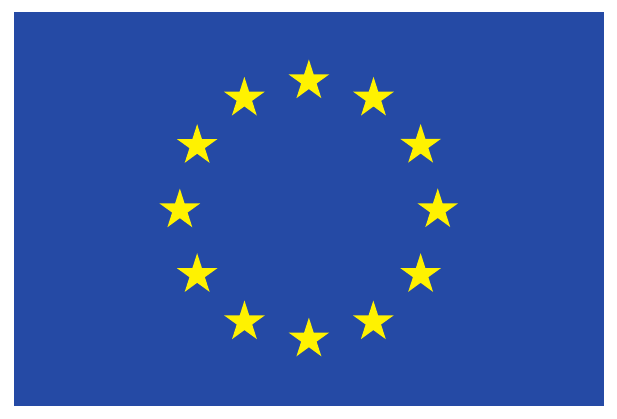}}
\end{minipage}
\hspace{0.01\linewidth}
\begin{minipage}{0.85\linewidth}
This work is part of the FoTran project, funded by the European Research Council (ERC) under the EU's Horizon 2020 research and innovation program (agreement \textnumero{}~771113). ~~We ~also ~thank ~the CSC-IT\vspace{0.3ex}
\end{minipage}
\begin{minipage}{\linewidth}
\noindent  Center for Science Ltd., for computational resources.
\end{minipage}
}

\bibliography{anthology,custom}

\appendix

\section{Corpus path templating}
\label{adx:pathtemplate}

Paths to corpora are specified using path templates, which can contain variables that will be substituted by the \texttt{config-config} tool.

\paragraph{Directional corpus mode.}
For corpora where the two translation directions of a language pair are distinguished from each other.

\begin{description}
\item[src\_lang] The source language of the task.
\item[tgt\_lang] The target language of the task.
\item[lang\_pair] \texttt{src\_lang-tgt\_lang} for convenience.
\end{description}

\paragraph{Symmetric corpus mode.}
For corpora where a language pair uses the same files for both translation directions.

\begin{description}
\item[lang\_a] The alphabetically first language.
\item[lang\_b] The alphabetically second language.
\item[side\_a] ‘src’ if the language pair is used in the “forward” direction, otherwise ‘trg’. Note that the abbreviation for target is ‘trg’, not ‘tgt’.
\item[side\_b] ‘trg’ if the language pair is used in the “forward” direction, otherwise ‘src’.
\item[sorted\_pair] the source and target languages in alphabetical order, separated by a hyphen.
\end{description}

As an example, let’s say that the corpus contains two files \texttt{eng-ben/train.src.gz} (English side) and \texttt{eng-ben/train.trg.gz} (Bengali side). The data should be used symmetrically for both Bengali-to-English and English-to-Bengali directions. For the first, \texttt{lang\_pair} and \texttt{sorted\_pair} are the same. For the second, \texttt{lang\_pair} is “eng-ben”, but \texttt{sorted\_pair} is “ben-eng”.

Thus, in order to use the files in the correct order, you should use the source path template \texttt{\{sorted\_pair\}/train.\{side\_a\}.gz}, and
\texttt{\{sorted\_pair\}/train.\{side\_b\}.gz}
as the target path template.

\section{Layerwise parameter sharing}
\label{adx:parametersharing}

The parameter sharing architecture is defined as two concatenations of modules, one for the encoder and one for the decoder. Each module has a specified number of layers, and a parameter sharing pattern. The following parameter sharing patterns are available in \texttt{config-config}. Note that arbitrary sharing patterns are possible when generating the configuration file by other means.

\begin{description}
\item[FULL] fully shared parameters. Will be named using the constant “full”.
\item[SRC\_GROUP] groupwise shared parameters. Will be named according to the cluster id of the source language.

\item[TGT\_GROUP] groupwise shared parameters. Will be named according to the cluster id of the target language.

\item[GROUP] groupwise shared parameters. Same as SRC\_GROUP for encoder and TGT\_GROUP for decoder. For convenience.

\item[SRC\_LANGUAGE] language specific parameters. Will be named according to the source language code.

\item[TGT\_LANGUAGE] language specific parameters. Will be named according to the target language code.

\item[LANGUAGE] language specific parameters. Same as SRC\_LANGUAGE for encoder and TGT\_LANGUAGE for decoder. For convenience.
\end{description}

Note that it is possible to have target-language-dependent modules in the encoder, by using TGT\_LANGUAGE or TGT\_GROUP in the definition of the encoder sharing patterns%
\footnote{Source-language dependent modules in the decoder are possible as well.}.

\section{Example meta-config}
\label{adx:metaconfig}

\begin{figure*}[ht!]
    \resizebox{0.475\linewidth}{!}{
        \includegraphics[trim={0.1cm 7.2cm 9.5cm 1cm},clip]{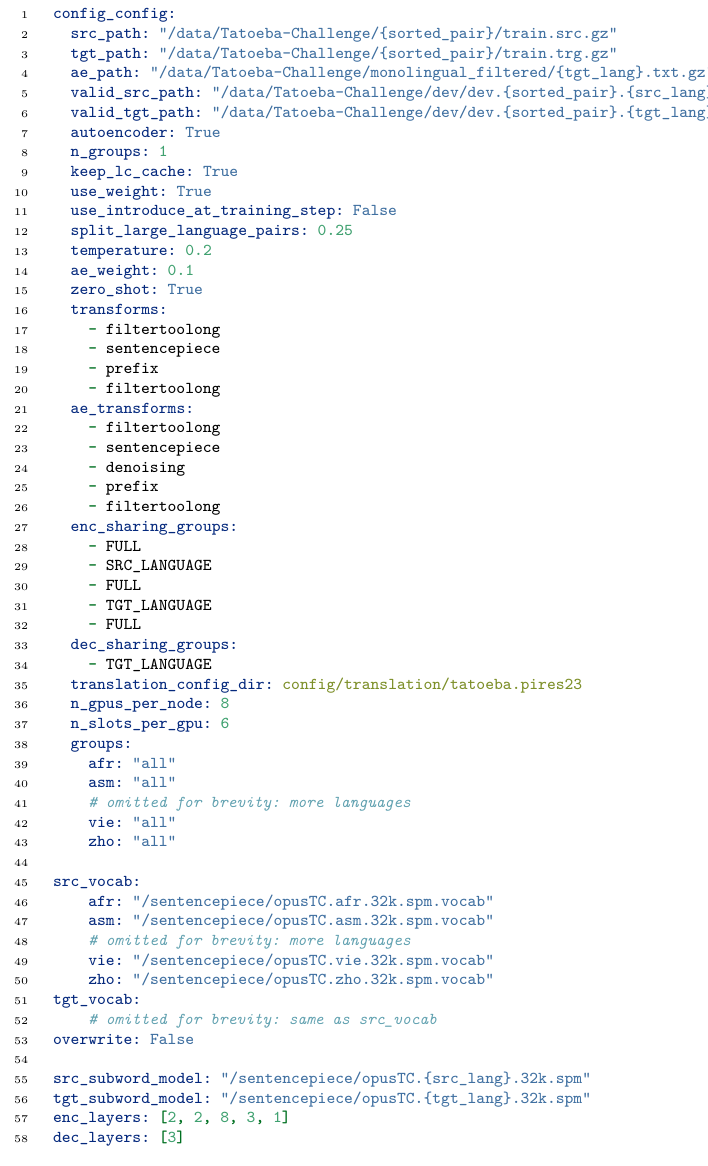}
    }
    \quad   
\resizebox{0.475\linewidth}{!}{
\includegraphics[trim={0.1cm 7.2cm 9.5cm 1cm},clip]{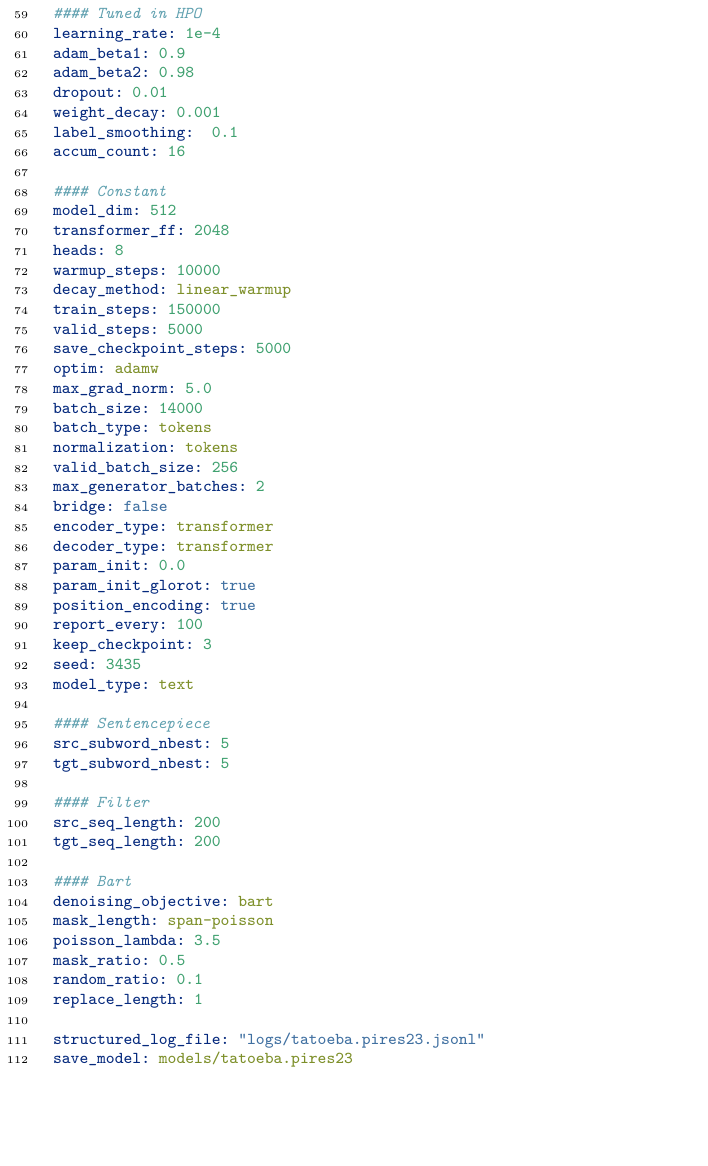}
}
\caption{Example meta-config}
\label{fig:metaconfig}
\end{figure*}

We include a practical example of a meta-configuration file in \Cref{fig:metaconfig}.

Note that the number of GPU devices per node (\texttt{n\_gpus\_per\_node}) and the maximum number of tasks to allocate per GPU (\texttt{n\_slots\_per\_gpu)} should be configured according to the cluster hardware.

\end{document}